\documentclass[letterpaper, 10 pt, conference]{ieeeconf}
\IEEEoverridecommandlockouts
\usepackage{cite}
\usepackage{amsmath,amssymb,amsfonts}
\usepackage{algorithm}
\usepackage{graphicx,import}
\usepackage{textcomp}
\usepackage{xcolor}
\usepackage{bm,nicefrac}
\usepackage{algpseudocode}
\usepackage{url}
\def\BibTeX{{\rm B\kern-.05em{\sc i\kern-.025em b}\kern-.08em
    T\kern-.1667em\lower.7ex\hbox{E}\kern-.125emX}}
    
\title{\LARGE \bf
A Novel Adaptive Controller for Robot Manipulators\\ Based on Active Inference
}

\author{Corrado Pezzato$^{1}$, Riccardo Ferrari$^{2}$ and Carlos Hern\'andez Corbato$^{3}$% <-this % stops a space
\thanks{*This research was supported by Ahold Delhaize. All content represents the opinion of the author(s), which is not necessarily shared or endorsed by their respective employers and/or sponsors.}% <-this % stops a space
\thanks{$^{1,3}$Corrado Pezzato and Carlos Hern\'andez Corbato are with the Cognitive Robotics Department,
        TU Delft, 2628 CD Delft, The Netherlands
        {\tt\small c.pezzato@tudelft.nl}, and {\tt\small c.h.corbato@tudelft.nl}} %
\thanks{$^{2}$Riccardo Ferrari is with the Department of Systems and Control, TU Delft, 2628 CD Delft, The Netherlands
        {\tt\footnotesize r.ferrari@tudelft.nl}}%
}

\begin{document}
\maketitle
\thispagestyle{empty}
\pagestyle{empty}

\begin{abstract}
More adaptive controllers for robot manipulators are needed, which can deal with large model uncertainties. This paper presents a novel active inference controller (AIC) as an adaptive control scheme for industrial robots. This scheme is easily scalable to high degrees-of-freedom, and it maintains high performance even in the presence of large unmodeled dynamics. The proposed method is based on active inference, a promising neuroscientific theory of the brain, which describes a biologically plausible algorithm for perception and action. In this work, we formulate active inference from a control perspective, deriving a model-free control law which is less sensitive to unmodeled dynamics. The performance and the adaptive properties of the algorithm are compared to a state-of-the-art model reference adaptive controller (MRAC) in an experimental setup with a real 7-DOF robot arm. The results showed that the AIC outperformed the MRAC in terms of adaptability, providing a more general control law. This confirmed the relevance of active inference for robot control. 
\end{abstract}

\section{Introduction}
\label{Sec:Introduction}
Traditional control approaches for industrial manipulators rely on an accurate model of the plant. However, there is an increasing demand in industry for robot controllers that are more flexible and adaptive to run-time variability. Often, robot manipulators are placed in dynamically changing surrounding, and they are subject to noisy sensory input and unexpected events.
In these new applications, obtaining such a model is a major problem. For example, in pick and place tasks, the dynamics of the robot manipulators can change unpredictably while handling unknown objects. Recent research has focused on the use of machine learning methods to obtain accurate inverse dynamic models \cite{NNgeneral1, NNgeneral2}.  In general, learning models using Neural Networks (NN) requires experts for defining the best topology for a particular problem \cite{NNMatteucci}.  Even though it is possible to exploit the physical knowledge of the system to simplify and improve the learning performance \cite{NNFOP}, the need of large amount of training data and several iterations for learning, still remains a problem and hard to generalise \cite{NNtask1, NNtask2}. Controllers that can dynamically adapt are required, but existing solutions in adaptive control either need an accurate model, or are difficult to tune and scale to higher number of DOFs.
In this paper, we present a novel adaptive controller for robot manipulators, inspired by a recent theory of the brain, which does not require accurate plant dynamics, and that is less sensitive to large parameters variation. 

The proposed control scheme is based on the general free-energy principle proposed by Karl Friston \cite{friston1}, and redefined in engineering terms \cite{buckley, tutorial}. The main idea at the basis of Friston's neuroscientific theory, is that the brain's cognition and motor control functions could be described in terms of energy minimization. It is supposed \cite{friston2} that humans have a set of sensory data and a specific internal model to characterize how the sensory data could have possibly been generated. Then, given this generative model, the causes of sensory data are inferred. Usually, the environment acts on humans to produce sensory impression, and humans can act on the environment to change it. In this view, the motor control of human body can be considered as the fulfillment of a prior expectation about proprioceptive sensations \cite{friston3}. Although the general active inference framework is mathematically well defined, its application to robotics remains a challenge. Active inference has mainly been applied to neuronal simulations (for handwriting \cite{friston1} for instance), supposing to know the true dynamical process. However, this is not the case in robotics. Even if the neuronal simulations are a strong proof of concept for the neuroscientific theory, in the present form their extension to realistic robotic scenarios \cite{lopez, arnau} does not provide advantages over other classical controllers. The main problems are the computational load and the definition of meaningful generative models. With our work we overcome these limitations, using active inference to derive a model-free control law. Instead of modeling the true unknown dynamical process, we define a reference model that active inference has to follow. The main contributions of this paper are twofold:
\begin{itemize}
    \item Derivation of an online active inference control law for a generic $n$-DOF robot manipulator in joint space.
    \item Comparison of the adaptability of the AIC with a state-of-the-art model reference adaptive controller.
\end{itemize}
The contributions have been experimentally validated in a 7-DOF collaborative industrial manipulator.

\subsection{Related work}
At present, the use of active inference for robot control is still limited. In \cite{lopez}, the authors simulated a PR2 robot controlled in Cartesian space for a reaching task. The solution was offline, computationally expensive, open-loop, and it relied on an additional position controller. This makes the approach not suitable for online tasks. A recent MSc thesis \cite{arnau}, based on \cite{lopez}, derived an offline closed-loop scheme of active inference. The feedforward torque commands for a simulated 7-DOF manipulator are computed offline, relying on additional controllers for feedback control. The scheme failed to control the robot in presence of gravity since the feedforward torques did not include the gravitational effect. Both \cite{lopez} and \cite{arnau} were based on the Statistical Parametric Mapping (SPM) by Friston. This toolbox is suitable for several offline applications, but it is too computationally heavy for online control. In \cite{arnau}, each iteration is reported to take about one second. Another recent work \cite{lanillos1} formalised the use of the free-energy for static state estimation, using a real UR5 robot arm equipped with proprioceptive and visual sensors. Even though the results of the state estimation were promising, no control actions were included. The same authors presented in \cite{lanillos2} the body estimation and control in the joint space of a simulated 2-DOF robot arm through active inference. This solution included state-of-the art regressors to estimate online the generative models. However, during the simulations, the estimation of the acceleration was unreliable and substituted with the ground truth. Regardless of the fact that only forward dynamics models had to be learned, the authors pointed out how this approach is not simpler compared with classical inverse dynamics techniques.  In a parallel, related work on active inference \cite{oliver}, the authors successfully controlled a real 3-DOF robot arm using velocity commands. In our approach we formulate an AIC for online closed loop control of industrial robots, using low-level torque commands. We also provide a comparison with a state-of-the-art adaptive controller, and insights for design and tuning. On the other hand, the adaptive control branch of control theory \cite{astrom}, offers solutions to deal with manipulators subject to parameters variation and abrupt changes in the dynamics. Within adaptive controllers, two main categories can be identified: the model reference adaptive systems, and the self-tuning regulators \cite{AdaptiveReview}. The first technique being studied for robot manipulators was the model reference adaptive control (MRAC) \cite{MRACreview}. The idea behind this technique is to derive a control signal to be applied to the robot actuators which will force the system to behave as specified by a chosen reference model. Furthermore, the adaptation law is designed to guarantee stability using either Lyapunov theory or hyperstability theory \cite{hyper}. The other most common approach for robot control is the self-tuning adaptive control \cite{selftuning1, selftuning2}. The main difference between this technique and the MRAC is that the self-tuning approach represents the robot as a linear discrete-time model and it estimates online the unknown parameters, substituting them in the control law. Adaptive control of robot manipulators is required in presence of uncertain dynamics and varying payloads, however, the complexity of the controller usually increases with increasing number of DOFs. Among all the possible adaptive controllers, in this paper we choose the MRAC with hyperstability theory \cite{hyper} for comparison. This approach provides adaptability to abrupt changes in the robot dynamics, and it does not require the kinematic or dynamic description of the manipulator, similarly to the AIC. 

\subsection{Paper structure}
The paper is organised as follows: In Sec. \ref{Sec:active-inference} we present the free-energy principle and active inference in control engineering terms. In Sec. \ref{Sec:AIC} we derive a novel AIC for a 7-DOF robot manipulator, and we explain the model assumptions and simplifications. In Sec \ref{Sec:MRAC} the MRAC is presented for comparison. In Sec. \ref{Sec:performance-analysis} we compare the adaptability of the AIC and MRAC in a simulated pick and place task, validating the results in the real setup. We also discuss the advantages of our AIC and the open questions. Finally, Sec.~\ref{Sec:Conclusions} provides a summary and directions for future work.

\section{The active inference framework}
\label{Sec:active-inference}
In this section we report the free-energy principle and active inference from \cite{friston2,buckley}, rewriting only the necessary concepts in control terms, to understand the derivation of our novel AIC in Sec. \ref{Sec:AIC}.
\subsection{The free-energy principle}
The free-energy principle is formulated in terms of Bayesian inference \cite{Bayes}. In this view, body perception for state estimation is framed using Bayes rule:
\begin{equation}
    p(\bm{x}|\bm{y}) = \frac{p(\bm{y}|\bm{x})p(\bm x)}{p(\bm y)}
\end{equation}
where $p(\bm{x}|\bm{y})$ is the probability of being in the $n$-dimensional state $\bm{x}$ given the current $m$-dimensional sensory input $\bm y$. Instead of exactly inferring the posterior, which often involves intractable integrals, an auxiliary probability distribution $r_d(\bm x)$, called recognition density, is introduced. By minimizing the Kullback-Leibler divergence ($D_{KL}$) between the true posterior $p(\bm{x}|\bm{y})$ and $r_d(\bm x)$, the most probable state given a sensory input is inferred \cite{buckley}. $D_{KL}$ is defined as:
\begin{equation}
    D_{KL}(r_d(\bm x)||p(\bm{x}|\bm{y})) = \int  r_d(\bm x) \ln\frac{r_d(\bm x)}{p(\bm{x}|\bm{y})}d\bm x = \mathcal{F}+\ln p(\bm y) 
\end{equation}
In the equation above, the scalar $\mathcal{F}$ is the so called \emph{free-energy}. By minimizing $\mathcal{F}$, $D_{KL}$ is also minimized and the recognition density approaches the true posterior. According to the Laplace approximation \cite{variational}, the controller only parametrises the sufficient statistics (e.g. mean and variance) of the recognition density. $r_d(\bm x)$ is assumed Gaussian and sharply peaked at its mean value $\bm \mu$. This approximation allows to simplify the expression for $\mathcal{F}$ which results:
\begin{equation}
\label{eq:F_prob}
    \mathcal{F}\approx -\ln p(\bm \mu,\bm y)
\end{equation}
The mean $\bm \mu$ is the internal belief about the true states $\bm x$. Minimizing $\mathcal{F}$, the controller is continuously adapting the internal belief $\bm \mu$ about the states $\bm x$ based on the current sensory input $\bm y$. 

\subsection{Free-energy equation}
Equation \eqref{eq:F_prob} is still general and it has to be further specified to numerically evaluate $\mathcal{F}$. To do so, the joint probability $p(\bm \mu,\bm y)$ has to be defined. This is done by introducing two generative models, one to predict the sensory data $\bm y$, according to the current belief $\bm \mu$, and another to describe the dynamics of the evolution of the belief $\bm \mu$.  
\subsubsection{Generative model of the sensory data}
The sensory data is modeled using the following expression \cite{buckley}:
\begin{equation}
    \label{eq:genMody}
    \bm y = \bm g(\bm \mu) +\bm z
\end{equation}
where $\bm g(\bm \mu)$ represents the non-linear mapping between sensory data and states of the environment, and $\bm z$ is Gaussian noise $\bm z \sim (\bm 0,\Sigma_y)$. The covariance matrix $\Sigma_y$ also represents the controller's confidence about each sensory input. 
\subsubsection{Generative model of the state dynamics}
In presence of time varying states $\bm x$, the controller has to encode a dynamic generative model of the evolution $\bm \mu'$ of the belief $\bm \mu$. This generative model is defined as\cite{buckley}:
\begin{equation}
    \label{eq:genModmu}
	\frac{d\bm{\mu}}{dt}=\bm \mu' = \bm{f}(\bm{\mu})+\bm{w}
\end{equation} 
where $\bm{f}$ is a generative function dependant on the belief about the states $\bm{\mu}$ and $\bm w$ is Gaussian noise $\bm w \sim (\bm 0,\Sigma_\mu)$.
\subsubsection{Generalised motions}
To describe the dynamics of the states, or better the belief about these dynamics, we have to introduce the concept of generalised motions \cite{friston5}. Generalised motions are used to represent the states of a dynamical system, using increasingly higher order derivatives of the states of the system itself. They apply to sensory inputs as well, meaning that the generalised motions of a position measurement, for example, correspond to its higher order temporal derivatives (velocity, acceleration, and so on). The use of generalised motions allows a more accurate description of the system's states. More precisely, the generalised motions $\bm{\tilde \mu}$ of the belief under local linearity assumptions \cite{variational} are, up to the second order:
\begin{eqnarray}
\label{genMotmu}
\nonumber
\bm{\mu} ' &=& \bm{\mu}^{(1)} =\bm{f}(\bm{\mu})+\bm{w}\\
\bm{\mu} '' &=& \bm{\mu}^{(2)} = \frac{\partial \bm{f}}{\partial \bm{\mu}}\bm{\mu} '+\bm{w}'
\end{eqnarray}
In general, we indicate the generalised motions of the states up to order $n_d$\footnote{Generalised motions can extend up to infinite order but the noise at high orders is predominant, thus we can limit the chosen order to $n_d$ \cite{friston4}.} as $\bm{\tilde{\mu}} = [\bm \mu,\ \bm \mu',\ \bm \mu'',\ \bm\mu''',\ ...,\  \bm\mu^{(n_d)}]$.\\
Similarly, the generalised motions of the sensory input are:
\begin{eqnarray}
\nonumber
\label{eq:genMoty}
\bm{y}  &=& \bm{y}^{(0)} = \bm{g}(\bm{\mu})+\bm{z}\\
\bm{y} '&=& \bm{y}^{(1)} = \frac{\partial \bm{g}}{\partial \bm{\mu}}\bm{\mu} '+\bm{z}'
\end{eqnarray}
We indicate the generalised motions of the sensory input up to order $n_d$ as $ \bm{\tilde{y}} = [\bm y,\ \bm y',\ \bm y'',\ \bm y''',\ ...,\  \bm y^{(n_d)}]$.
\subsubsection{General free-energy expression} 
With the extra theoretical knowledge about the generalised motions, we can define an expression for the free-energy for a multivariate case in a dynamically changing environment:
\begin{equation}
    \label{eq:FlogAIC}
    \mathcal{F} = -\ln{p(\tilde{\bm \mu},\tilde{\bm y})}
\end{equation}
%The likelihood of the sensory data $p(\tilde{\bm y}|\tilde{\bm \mu})$ and the prior $p(\tilde{\bm \mu})$ have to be specified. To do so, a few considerations have to be made. 
The joint probability $p(\tilde{\bm \mu},\tilde{\bm y})$ has to be specified. According to \cite{buckley} and to the definitions previously given, the noise at each dynamical order is considered uncorrelated. Then, according to the generalised sensory input, the sensory data at a particular order relates only with the states at the same dynamical order. Similarly, for the state dynamics, the state at a certain dynamical order are related only with those which are one order below. Then, using the chain rule, it results:
\begin{equation}
\label{eq:p_mutildeytilde}
    p(\tilde{\bm \mu},\tilde{\bm y}) = \prod_{i=0}^{n_d-1}p(\bm y^{(i)}|\bm \mu^{(i)})p(\bm \mu^{(i+1)}|\bm \mu^{(i)})
\end{equation}
Using the Laplace assumption, and thus considering Gaussian distributed probability densities, we can write:
\begin{eqnarray}
    \nonumber
    \label{eq:p(mu+|mu)p(y|mu)}
    &p(\bm \mu^{(i+1)}|\bm\mu^{(i)})= \frac{1}{|\Sigma_{\mu^{(i)}}|\sqrt[n]{2\pi}}\exp\left\{-\frac{1}{2}\bm \varepsilon_\mu^{(i)\top}\Sigma^{-1}_{\mu^{(i)}}\bm \varepsilon_\mu^{(i)}\right\}\\
    &p(\bm y^{(i)}|\bm \mu^{(i)})= \frac{1}{|\Sigma_{y^{(i)}}|\sqrt[n]{2\pi}}exp\left\{-\frac{1}{2}\bm \varepsilon_y^{(i)\top}\Sigma^{-1}_{y^{(i)}}\bm \varepsilon_y^{(i)}\right\}
\end{eqnarray}
where $\bm \varepsilon_y^{(i)}=(\bm{y}^{(i)}-\bm{g}^{(i)}(\bm{\mu}))$ and $\bm \varepsilon_\mu^{(i)}=(\bm{\mu}^{(i+1)}-\bm{f}^{(i)}(\bm \mu))$ are respectively the sensory and state model prediction errors. Furthermore it holds:
\begin{equation}
\bm{g}^{(i)}=\frac{\partial \bm{g}}{\partial \bm{\mu}}\bm{\mu}^{(i)}, \hspace{2mm} \bm{f}^{(i)}=\frac{\partial \bm{f}}{\partial \bm{\mu}}\bm{\mu}^{(i)}, \hspace{2mm} \bm{g}^{(0)}= \bm g, \hspace{2mm} \bm{f}^{(0)}= \bm f
\end{equation}
Substituting \eqref{eq:p_mutildeytilde} in \eqref{eq:FlogAIC} leads to:
\begin{equation}
    \label{eq:FSum}
    \mathcal{F} = -\sum_{i=0}^{n_d-1}\left[\ln{p(\bm y^{(i)}|\bm \mu^{(i)})}+\ln{p(\bm \mu^{(i+1)}|\bm \mu^{(i)})}\right]
\end{equation}
Finally, according to \eqref{eq:p(mu+|mu)p(y|mu)}, $\mathcal{F}$ can be expressed up to a constant as a weighted sum of squared prediction errors:
\begin{equation}
    \label{eq:F}
	\mathcal{F} = \frac{1}{2}\sum_{i=0}^{n_d-1}\left[ \bm \varepsilon_y^{(i)\top}\Sigma^{-1}_{y^{(i)}}\bm \varepsilon_y^{(i)}+\bm \varepsilon_\mu^{(i)\top}\Sigma^{-1}_{\mu^{(i)}}\bm \varepsilon_\mu^{(i)}\right]+K
\end{equation}
where $n_d$ is the number of generalised motions chosen and $K$ is a constant term resulting from the substitution. The minimisation of this expression can be done by refining the internal belief, thus performing state estimation, but also computing the control actions to fulfill the prior expectations and achieve a desired motion. The constant term $K$ is neglected in the sequel since it plays no role into the minimisation problem. The next two subsections describe the approach proposed by Friston \cite{friston2,friston6} to minimise $\mathcal{F}$, using gradient descent. 

\subsection{Belief update for state estimation}
The belief update law for state estimation is determined from the gradient of the free-energy, with respect to each generalised motion \cite{buckley, friston5}:
\begin{equation}
    \label{eq:state_update_general}
        \dot{\bm{\tilde{\mu}}} = \frac{d}{dt}\tilde{\bm \mu} -\kappa_\mu\frac{\partial \mathcal{F}}{\partial \bm{\tilde{\mu}}}
\end{equation}
The learning rate $\kappa_{\mu}$, can be seen from a control perspective as a tuning parameter for the state update. 
\subsection{Control actions}
In the free-energy principle the control actions play a fundamental role in the minimisation process. In fact, the control input $\bm u$ allows to steer the system to a desired state while minimising the prediction errors in $\mathcal{F}$. This is done using gradient descent. Since the free-energy is not a function of the control actions directly, but the actions $\bm u$ can influence $\mathcal{F}$ by modifying the sensory input, we can write \cite{buckley}:
\begin{equation}
    \frac{\partial \mathcal{F}(\bm{\tilde{\mu}},\bm{\tilde{y}}(\bm{u}) )}{\partial \bm{u}}=\frac{\partial \bm{\tilde{y}}(\bm{u})}{\partial \bm{u}}\frac{\partial \mathcal{F}(\bm{\tilde{\mu}},\bm{y}(\bm{u}) )}{\partial \bm{\tilde{y}}(\bm{u})}
\end{equation}
Dropping the dependencies for a more compact notation, the dynamics of the control actions can be written as:
\begin{equation}
    \label{eq:actions_general}
	\dot{\bm{u}}=-\kappa_a\frac{\partial \bm{\tilde{y}}}{\partial \bm{u}}\frac{\partial \mathcal{F}}{\partial \bm{\tilde{y}}}
\end{equation}
where $\kappa_a$ is the tuning parameter to be chosen. 
\section{Robot arm control with active inference}
\label{Sec:AIC}
In this section we derive the first model-free, computationally lightweight, online torque controller for joint space control using active inference. The established theory of Sec.~\ref{Sec:active-inference} is adapted to define a novel control scheme for a generic $n$-DOF manipulator. The challenging problem of finding suitable generative models $\bm f(\cdot)$ and $\bm g(\cdot)$, and the relation $\partial \bm{\tilde{y}}/\partial \bm{u}$ in such a complex scenario is solved.
\subsubsection*{Assumption 1}  The robot manipulator is equipped with position and velocity sensors, which respectively provide the two variables $\bm y_q,\ \bm y_{\dot{q}} \in \mathbb{R}^n$.
\subsubsection*{Assumption 2} Since only the position and velocity measurements are available, we will consider the generalised motions up to order two, so $n_d=2$. %Doing so, \eqref{genMotmu} and \eqref{eq:genMoty} reduce to:
%\begin{equation}
%\label{eq:genMot_simplified}
%\begin{cases}
%	\bm{\mu} ' & = \bm{f}(\bm{\mu})+\bm{w}\\
%	\bm{\mu} '' & = \frac{\partial \bm{f}}{\partial \bm{\mu}}\bm{\mu} '+\bm{w}'\\
%\end{cases} \hspace{5mm}
%\begin{cases}
%	 \bm{y}_q & = \bm{g}(\bm{\mu})+\bm{z}\\
%     \bm y_{\dot{q}} & = \frac{\partial \bm{g}}{\partial \bm{\mu}} \bm{\mu}'+\bm{z}'\\
%\end{cases}
%\end{equation} 
\subsubsection*{Assumption 3} The Gaussian noise affecting the different sensory channels is supposed uncorrelated \cite{variational, buckley}. The covariance matrices for sensory input and state belief are:
\begin{eqnarray}
\label{eq:bigsigmas}
    \Sigma_{y^{(0)}} = \sigma_q I_n,\   \Sigma_{y^{(1)}} = \sigma_{\dot{q}} I_n,\\
    \Sigma_{\mu^{(0)}} = \sigma_\mu I_n,\  \Sigma_{\mu^{(1)}} = \sigma_{\mu'}I_n
\end{eqnarray}
where we supposed that the controller associates four different variances to describe its confidence about sensory input and internal belief. 
\subsubsection*{Assumption 4} The states of the environment $\bm x$ are set as the joint positions of the robot arm. Doing so, we can control the robot arm in joint space through free-energy minimization, and simplify the equations for states update and control actions.
\subsection{Generative models and $\mathcal{F}$ for a robot manipulator}
In order to numerically evaluate the free-energy as in \eqref{eq:F}, the two functions $\bm g(\bm \mu)$ and $\bm f(\bm \mu)$ have to be chosen. 
\subsubsection{Generative model of the sensory data}
$\bm g(\bm \mu)$ indicates the relation between the sensed values and the states. Since we chose the states to be the joint positions and the sensory data provides directly the noisy values $\bm y_q$ and $\bm y_{\dot{\bm q}}$, it holds:
\begin{equation}
    \label{eq:g_mu}
    \bm{g}_q(\bm{\mu}) = \bm{\mu}, \hspace{10mm} \partial \bm{g}_q/\partial \bm{\mu} = 1
\end{equation}
\subsubsection{Dynamic generative model of the world} Instead of modelling the true dynamics of the manipulator, we propose to define a reference model to specify the desired behaviour of the robot \cite{buckley}. In particular, the world dynamics are chosen such that the robot is steered to a desired position $\bm \mu_d$. In other words, the controller believes that the states will evolve in such a way that they will reach the goal $\bm \mu_d$ with the dynamics of a first order system with unitary time constant: 
\begin{equation}
	\bm{f}(\bm{\mu}) = \bm{\mu}_d-\bm{\mu}
	\label{eq:f_mu}
\end{equation}
The value $\bm \mu_d$ is a constant $\in \mathbb{R}^n$ corresponding to the desired set-point for the joints of the manipulator. Substituting \eqref{eq:g_mu} and \eqref{eq:f_mu} in \eqref{eq:genMoty} and  \eqref{genMotmu}, it results:
\begin{equation}
\begin{cases}
	\bm{\mu} ' & = \bm{\mu}_d-\bm{\mu}+\bm{w}\\
	\bm{\mu} '' & = -\bm{\mu} '+\bm{w}'\\
\end{cases} \hspace{10mm}
\begin{cases}
	\bm y_q & = \bm{\mu}+\bm{z}\\
	\bm y_{\dot{q}} & =  \bm{\mu}'+\bm{z}'\\
\end{cases}
\label{eq:genMot_final}
\end{equation} 
According to \eqref{eq:genMot_final} and \eqref{eq:F}, the free-energy expression for a generic robot manipulator under the assumptions given is:
\begin{eqnarray}
	\label{eq:F_manipulator}
    \nonumber
	\mathcal{F} &=& \frac{1}{2}(\bm{y}_q-\bm{\mu})^\top\Sigma_{y^{(0)}}^{-1}(\bm{y}_q-\bm{\mu})\\
	\nonumber
	&+&\frac{1}{2}(\bm y_{\dot{q}}-\bm{\mu}')^\top\Sigma_{y^{(1)}}^{-1}(\bm y_{\dot{q}}-\bm{\mu}')\\
    \nonumber
	&+& \frac{1}{2}(\bm{\mu}'+\bm{\mu}-\bm{\mu}_d)^\top\Sigma_{\mu^{(0)}}^{-1}(\bm{\mu}'+\bm{\mu}-\bm{\mu}_d)\\
	&+&\frac{1}{2}(\bm{\mu}''+\bm{\mu}')^\top\Sigma_{\mu^{(1)}}^{-1}(\bm{\mu}''+\bm{\mu}')
\end{eqnarray}
\subsection{Belief update and state estimation for a manipulator}
According to the free-energy principle, the states of the robot manipulator can be estimated using a gradient descent scheme. Applying \eqref{eq:state_update_general}, having defined $\mathcal{F}$ as in \eqref{eq:F_manipulator}, leads to the following state update law:
\begin{eqnarray}
	\label{mudot_manipulator}
	\nonumber
	\dot{\bm{\mu}}  &=&	\bm{\mu}'+\kappa_\mu\Sigma_{y^{(0)}}^{-1}(\bm{y}_q-\bm{\mu})-\kappa_\mu\Sigma_{\mu^{(0)}}^{-1}(\bm{\mu}'+\bm{\mu}-\bm{\mu}_d)\\
	\nonumber
	\dot{\bm{\mu}}'	&=& \bm{\mu}''+\kappa_\mu\Sigma_{y^{(1)}}^{-1}(\bm y_{\dot{q}}-\bm{\mu}')-\kappa_\mu\Sigma_{\mu^{(0)}}^{-1}(\bm{\mu}'+\bm{\mu}-\bm{\mu}_d)\\
	\nonumber
	&-&\kappa_\mu\Sigma_{\mu^{(1)}}^{-1}(\bm{\mu}''+\bm{\mu}')\\
	\dot{\bm{\mu}}''&=& -\kappa_\mu\Sigma_{\mu^{(1)}}^{-1}(\bm{\mu}''+\bm{\mu}')
\end{eqnarray}
Note that $\kappa_\mu$ is the tuning parameter for state estimation.
\subsection{Control actions for a robot manipulator}
The final step in order to be able to steer the joints of a robot manipulator to a desired value $\bm \mu_d$, is the definition of the control actions. 
\subsubsection{General considerations}
The general actions update is expressed by \eqref{eq:actions_general}. The partial derivatives of \eqref{eq:F_manipulator} with respect to the generalised sensory input are given by:
\begin{equation}
	\frac{\partial \mathcal{F}}{\partial \bm{y}_q} = \Sigma_{y^{(0)}}^{-1}(\bm{y}_q-\bm{\mu}),\hspace{5mm} 
	\frac{\partial \mathcal{F}}{\partial \bm y_{\dot{q}}} = \Sigma_{y^{(1)}}^{-1}(\bm y_{\dot{q}}-\bm{\mu}')
\end{equation} 
Having said that, the actions update is expressed as:
\begin{eqnarray}
	\label{eq:u_Full}
	\dot{\bm{u}}=-\kappa_a\begin{bmatrix}
	\frac{\partial \bm y_q}{\partial \bm u}\Sigma_{y^{(0)}}^{-1}(\bm{y}_q-\bm{\mu})+\frac{\partial \bm y_{\dot{q}}}{\partial \bm u}\Sigma_{y^{(1)}}^{-1}(\bm y_{\dot{q}}-\bm{\mu}')\end{bmatrix}
\end{eqnarray}
Active inference requires then to define the change in the sensory input with respect to the control actions, namely $\nicefrac{\partial \bm y_q}{\partial \bm u}$ and $\nicefrac{\partial \bm y_{\dot{q}}}{\partial \bm u}$. This is usually a hard forward dynamic problem, which constituted a major complication in past control strategies. One approach to compute these relations is through online learning using high-dimensional space regressors. However, this increases the complexity of the overall scheme and can produce unreliable results, as shown by the authors in \cite{lanillos2}. In this paper we propose to approximate the partial derivatives relying on the high adaptability of the active inference controller against unmodeled dynamics, as suggested in the conclusive remarks in \cite{lanillos2}. 
\subsubsection{Approximation of the true relation between $\bm u$ and $\bm{\tilde y}$} Let us first analyse the structure of the partial derivative matrices in \eqref{eq:u_Full}. The control action is a vector of $n$ torques applied to the $n$ joints of the robot manipulator. Each torque has a direct effect only on the corresponding joint to which it is applied. This allows us to conclude that $\nicefrac{\partial \bm y_q}{\partial \bm u}$ and $\nicefrac{\partial \bm y_{\dot{q}}}{\partial \bm u}$ are diagonal matrices.  Furthermore, considering the second Newton's law, the total torque applied to a rotational joint equals the moment of inertia times the angular acceleration. The diagonal terms of the partial derivatives matrices are then time varying positive values which depend on the current robot configuration. In other words, this means that a positive torque applied to a joint will always result in a positive contribution for both position and velocity of that specific joint. In this control scheme we propose to approximate the true positive time-varying relation with a positive constant, making use of the learning rate $\kappa_a$ as tuning parameter to achieve a sufficiently fast actions update. The control update law is finally given by:
\begin{equation}
	\label{eq:u_2DOFFull}
	\dot{\bm{u}}=-\kappa_a\begin{bmatrix}
	C_q\Sigma_{y^{(0)}}^{-1}(\bm{y}_q-\bm{\mu})+C_{\dot{q}}\Sigma_{y^{(1)}}^{-1}(\bm y_{\dot{q}}-\bm{\mu}')
	\end{bmatrix}
\end{equation}
\begin{equation}
\label{eq:partQ}
	 \frac{\partial \bm y_q}{\partial \bm u} \approx C_q,\
	 \frac{\partial \bm y_{\dot{q}}}{\partial \bm u} \approx C_{\dot{q}}
\end{equation}
The positive definite diagonal constant matrices $C_q,\ C_{\dot{q}}$ are then set to the identity, meaning that we only encode the sign of the relation between $\bm u$ and the change in $\bm{\tilde y}$.
\subsubsection{Tuning parameters AIC} The tuning parameters for the active inference controller are: 
\begin{itemize}
    \item $\sigma_q,\ \sigma_{\dot{q}} ,\  \sigma_\mu,\  \sigma_{\mu'}$: the standard deviations representing the confidence of the controller regarding its sensory input and internal belief about the states;
    \item $\kappa_\mu$, $\kappa_a$: the learning rates for state update and control actions respectively.
\end{itemize}
Algorithm \ref{alg:AIC} reports the pseudo-code of our AIC. For state and actions update, first-order Euler integration is used.
\begin{algorithm}
\caption{AIC for robot control}\label{alg:AIC}
\begin{algorithmic}[0]
\State \textbf{\textit{{Initialization}}}
\State $Par \gets \sigma_q,\ \sigma_{\dot{q}} ,\  \sigma_\mu,\  \sigma_{\mu'},\ \kappa_\mu$, $\kappa_a$\Comment{Set AIC parameters}
\State $\bm \mu = \bm y_q\in \mathbb{R}^n$\Comment{Initialise belief}
\State $\bm \mu'= \bm y_{\dot{q}}\in \mathbb{R}^n$
\State $\bm \mu'' = \bm 0\in \mathbb{R}^n$
\State $\bm u = \bm 0\in \mathbb{R}^n$\Comment{Initialise torque commands}
\State $\bm \mu_d \in \mathbb{R}^n$\Comment{Set prior, desired goal}
    \\\hrulefill
\State \textbf{\textit{{Control Loop}}}\Comment{At high frequency}
\State $\bm y_q,\ \bm y_{\dot{q}}$\Comment{Retrieve sensory input}
\State $\dot{\bm{\tilde{\mu}}} = \frac{d}{dt}\tilde{\bm \mu} -\kappa_\mu\frac{\partial \mathcal{F}}{\partial \bm{\tilde{\mu}}}$ \Comment{Belief dynamics \eqref{eq:state_update_general}}
\State $\bm{\tilde{\mu}}= \bm{\tilde{\mu}}+\Delta_t\dot{\bm{\tilde{\mu}}}$\Comment{Belief update, integration}
\State $\dot{\bm{u}}=-\kappa_a\frac{\partial \bm{\tilde{y}}}{\partial \bm{u}}\frac{\partial \mathcal{F}}{\partial \bm{\tilde{y}}}$\Comment{Action dynamics \eqref{eq:actions_general}}
\State $\bm{u}= \bm{u}+\Delta_t\dot{\bm{u}}$\Comment{Action update, integration}
\State \textbf{return} $\bm u$\Comment{Commanded torque}
\end{algorithmic}
\end{algorithm}

\section{Model reference adaptive controller}
\label{Sec:MRAC}
The controller chosen for comparison is an MRAC. This adaptive controller allows to obtain decoupled joint dynamics, forcing every single joint $i=1,...,n$ to respond as a second order linear system with transfer function:
\begin{equation}
	G_i(s)=\frac{\omega_i^2}{s^2+2\zeta\omega_is+\omega_i^2}q_{ri}(s)
	\label{eq:secondordermodel}
\end{equation}
The control architecture is taken from \cite{hyper}, where the control is specified in terms of feedforward and feedback adaptive gain matrices. These time-varying gain matrices are adjusted by means of adaptation laws to guarantee closed loop stability in case of large parameters perturbations. Supposing zero initial conditions for the gains, and neglecting the derivative terms as described in \cite{hyper}, it holds:
\begin{eqnarray}
	K_0(t)&=& E_{01}\bar{\bm q}_e(t)\bm q(t)^\top+E_{02}\int_0^T{\bar{\bm q}_e(\tau)q(\tau)d\tau}\\
	K_1(t)&=&E_{11}\bar{\bm q}_e(t)\dot{\bm q}(t)^\top+E_{12}\int_0^T{\bar{\bm q}_e(\tau)\dot{\bm q}(\tau)d\tau}\\
	Q_0(t)&=&F_{01}\bar{\bm q}_e(t)\bm q_r(t)^\top+F_{02}\int_0^T{\bar{\bm q}_e(\tau)\bm q_r(\tau)d\tau}\hspace{6mm}\\
	Q_1(t)&=&F_{11}\bar{\bm q}_e(t)\dot{\bm q}_r(t)^\top+F_{12}\int_0^T{\bar{\bm q}_e(\tau)\dot{\bm q}_r(\tau)d\tau}\\
	f(t)&=&\alpha_1\bar{\bm q}_e(t)+\alpha_2\int_0^T{\bar{\bm q}_e(\tau)d\tau}
	\label{eq:adaptlmatrices}
\end{eqnarray}
The variables $\bm q_r$ and $\dot{\bm q}_r$ are the desired references to track. The diagonal matrices $E_{jk}$ and $F_{jk}$ $\in \mathbb{R}^{n\times n}$, and the vector $\alpha_k$ $\in \mathbb{R}^{n}$ with $j=\{0,1\}$ and $k=\{1,2\}$, are the tuning parameters for the proportional-integral adaptation law. 
The term $\bar{\bm q}_e$ is called modified joint angle error vector \cite{hyper}:
\begin{equation}
	\bar{\bm q}_e = P_2[\bm q_r(t)-\bm q(t)]+P3[\dot{\bm q}_r(t)-\dot{\bm q}(t)]
\end{equation}
with $P2$ and $P3$ diagonal weighting matrices. The MRAC, similarly to the AIC, does not need the dynamic description of the robot manipulator, and it is scalable to high DOF. However, the number of the tuning parameters increases with the degrees of freedom, unlike for the AIC. 

\section{Experimental Evaluation}
\label{Sec:performance-analysis}
\begin{figure}[!htb]
    \centering
    \includegraphics[width=0.42\textwidth]{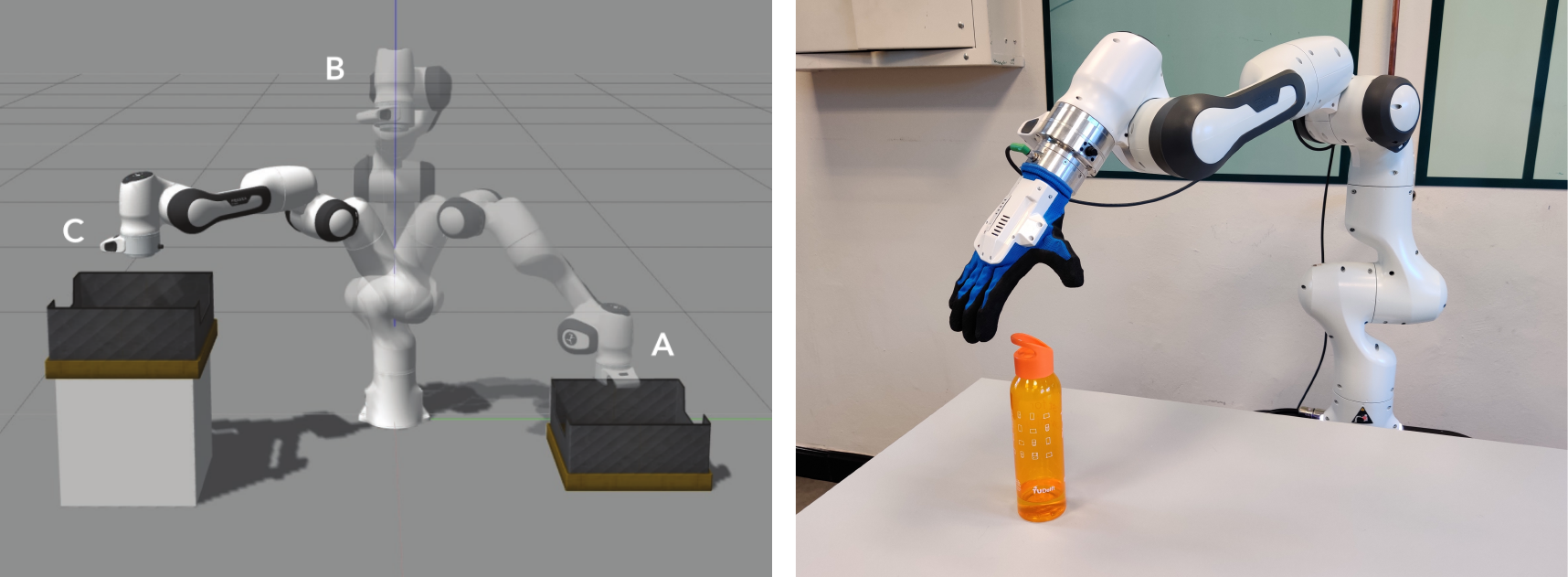}
    \caption{Simulated and real robot for pick and place cycle. }
    \label{fig:pandaGazebo}
\end{figure}

The adaptive properties of AIC and MRAC are now compared: The controllers are tuned in simulation using an approximated model of the robot, and then transferred to the real system.  The tests performed are based on a pick and place cycle using the Franka Emika Panda~7-DOF robot manipulator, as in Fig. \ref{fig:pandaGazebo}, with different payloads.
\subsection{Remarks about the tuning procedure for the controllers}
Before presenting the simulations and experimental results, we provide some observations regarding the number of parameters and the different tuning procedures for the AIC and MRAC.
\subsubsection{Number of tuning parameters}
The number of tuning parameters for the MRAC equals the number of DOFs times the number of weighting terms. According to Sec. \ref{Sec:MRAC}, this results in $17\times n$ parameters to be tuned. Regarding the AIC, instead, the number of tuning parameters is independent from the DOFs and it equals 6, following the formulation presented in Sec. \ref{Sec:AIC}. The lower number of parameters resulted in an overall easier tuning procedure for the active inference controller. 
As a final remark, to modify the behaviour of the step response for the AIC, such as rise time and settling time, one should change the internal reference model $\bm{f}(\bm \mu)$ instead of fine tuning the controller's parameters. 

\subsubsection{AIC tuning procedure}
To obtain a satisfactory response for the AIC, we performed the following steps: 1)~We set the controller confidence about sensory input and internal belief to one; 2)~We disabled the control actions and incremented the learning rate $\kappa_\mu$ until the state estimation in a static situation was fast enough; 3)  We included the control actions and increased the learning rate $\kappa_a$ until the robot was steered to the desired position, showing significant oscillations; 4) We dampened the oscillatory behaviour decreasing the sensory confidence about the most noisy sensors and the internal belief about velocities. 
\subsection{Simulations with approximated model}
\begin{figure}[!htb]
    \centering
    \includegraphics[width=0.45\textwidth]{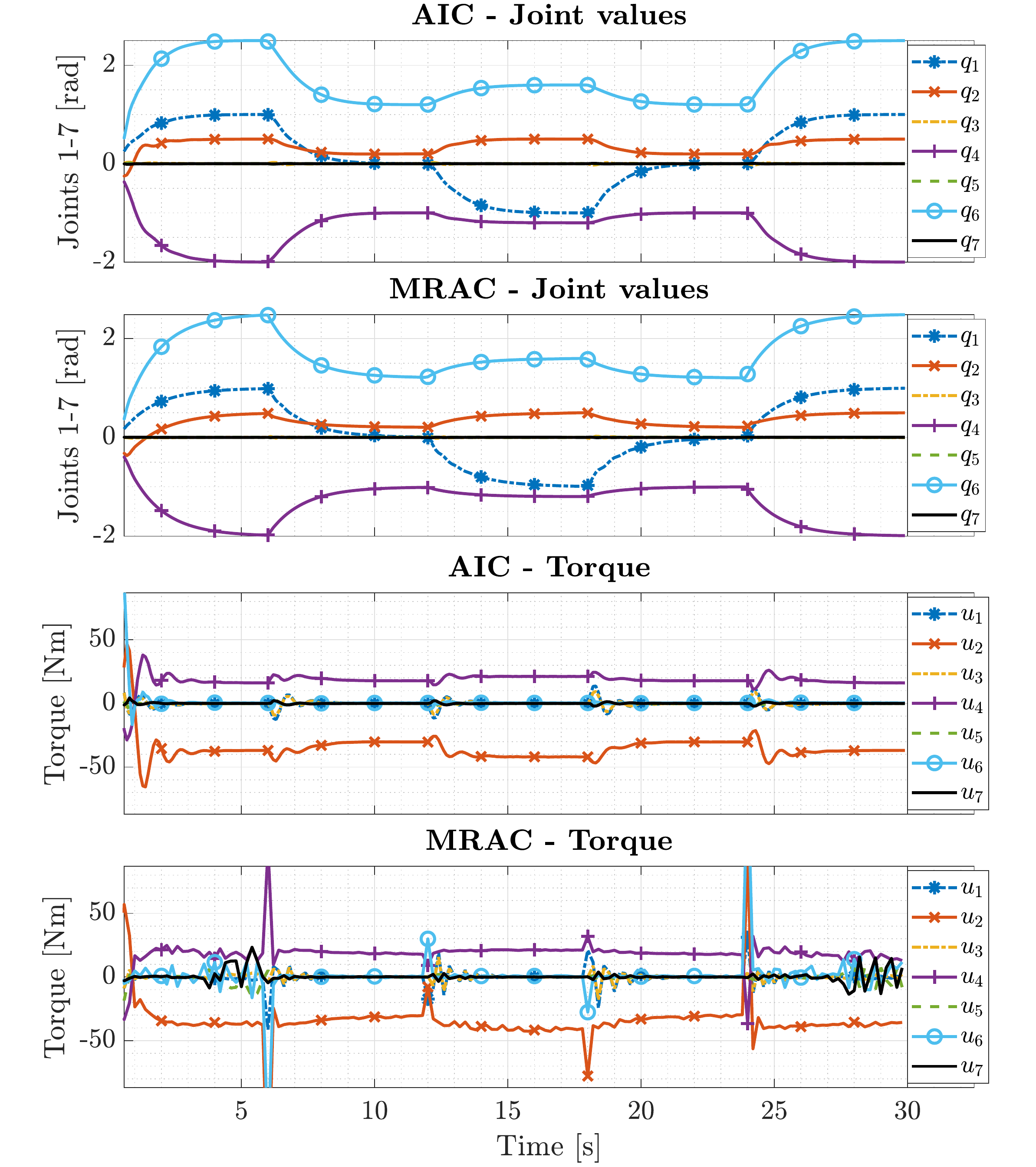}
    \caption{Response and control actions for the 7-DOF robot arm controlled through AIC and MRAC with approximated dynamics.}
    \label{fig:AIC_MRAC_model}
\end{figure}

The performance of AIC and MRAC in simulation are now presented. The task is a pick and place cycle where the desired joint values are chosen such that the arm simulates the pick and place of an object from one bin to the other, positioning the end-effector in A, B or C, see Fig \ref{fig:pandaGazebo}. This is achieved giving every $6\ [s]$ a set-point in joint space following the sequence: $\bm q_A$, $\bm q_B$, $\bm q_C$, $\bm q_B$, $\bm q_A$, where:
\begin{itemize}
    \item $\bm q_A=[1,\ 0.5,\ 0,\ -2,\ 0,\ 2.5,\ 0]\ [rad]$
    \item $\bm q_B=[0,\ 0.2,\ 0,\ -1,\ 0,\ 1.2,\ 0]\ [rad]$
    \item $\bm q_C=[-1,\ 0.5,\ 0,\ -1.2,\ 0,\ 1.6,\ 0]\ [rad]$
\end{itemize}
The controllers have been tuned using a considerably inaccurate model of the robot arm on purpose. The links have been approximated as cuboids, and 20\% random uncertainty in each link's mass has been assumed. This will allow to evaluate later on the adaptability performance while applying the controllers to the real manipulator. The joint values and control actions using AIC and MRAC, are depicted in Fig.~\ref{fig:AIC_MRAC_model}. Note that, for the MRAC, saturation of the control input at $\pm 85Nm$ is reached for some of the joints, after providing the new goal position. 
\begin{figure*}
    \centering
    \includegraphics[width=0.9\textwidth]{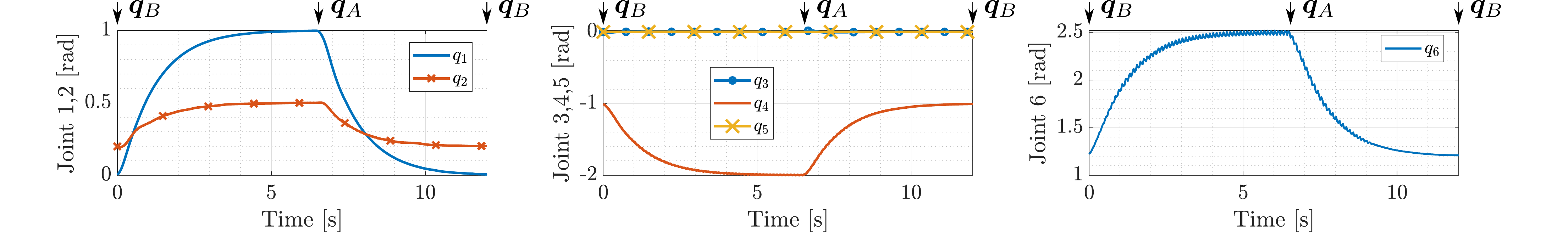}
    \caption{AIC on real setup without re-tuning from simulation. Focus on initial part of the pick and place cycle ($\bm q_B \to \bm q_A \to \bm q_B$) to highlight jittering}
    \label{fig:experiments}
\end{figure*}
\begin{figure*}
    \centering
    \includegraphics[width=0.9\textwidth]{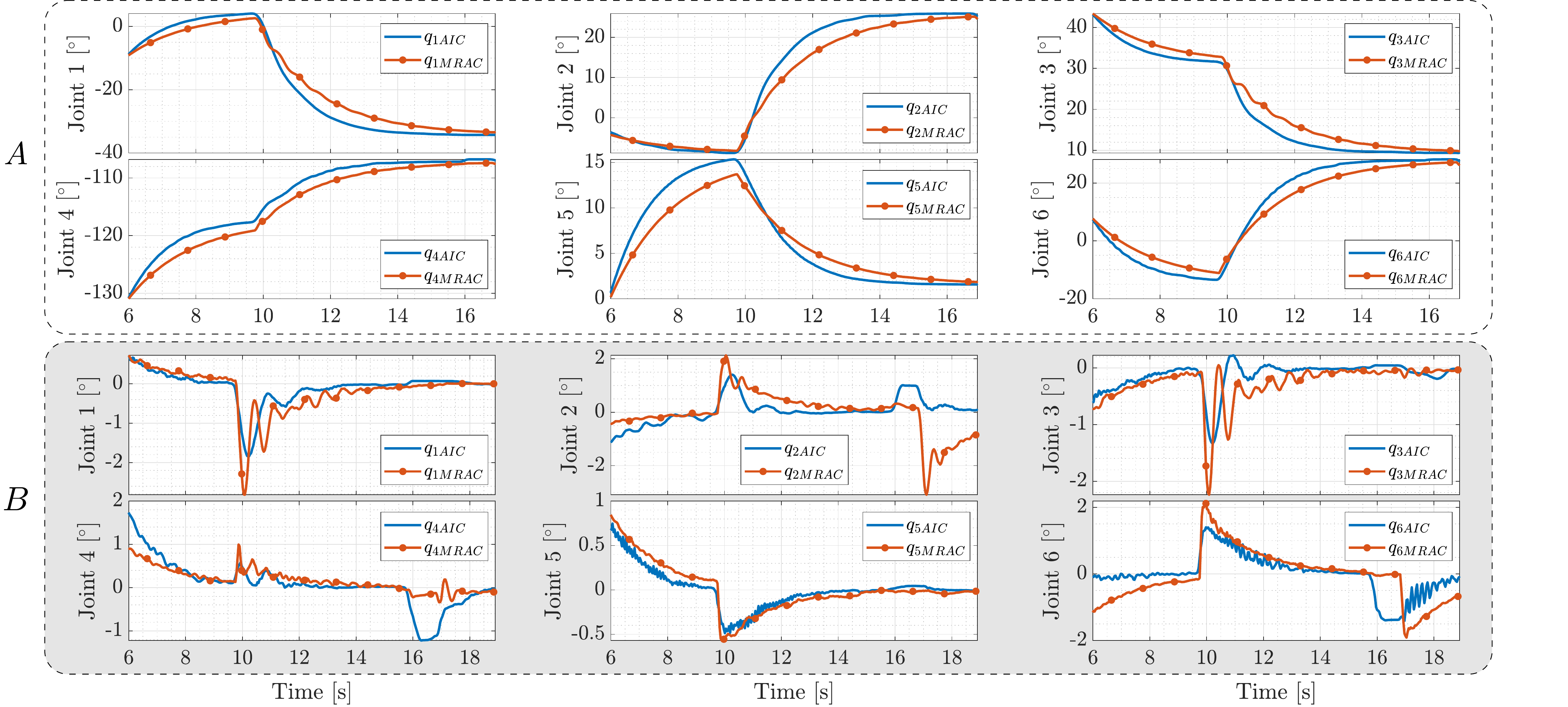}
    \caption{(A) Lift and place of empty bottle. (B) Difference of trajectories between empty and filled bottle during lift, place, and release}
    \label{fig:pickplacesimreal}
\end{figure*}
\subsection{Experiments on the real setup}
The same controllers tuned in simulation using the approximated model of the 7-DOF robot arm are now applied to control the real manipulator. Two tests are performed: first, the pick and place cycle of the previous section is repeated in the real robot, without re-tuning the controllers. Second, the AIC and MRAC are re-tuned in the real robot and used to pick and place different objects.  The real setup is controlled using a standard laptop running Ubuntu 16.04 with RT kernel, 8-cores Intel i7-4710MQ 2.50GHz.

\subsubsection{Pick and place cycle on the real robot}
We applied the MRAC and AIC from simulation to the real 7-DOF Franka Emika Panda. It is important to notice that, besides having different physical parameters, the real setup is already gravity compensated. The AIC and MRAC are simply applied on top of this intrinsic controller. This is already a considerable change in the system's dynamics, but to further increase the level of uncertainties, an end-effector is attached to the robot. From a modeling point of view, the system used for tuning the controllers in simulation is completely different from the real one. Usually, a controller tuned in simulation will not directly work on a real setup, especially if the initial model was not accurate. This was indeed the case for the MRAC which, when transferred to the real robot, could not control the setup leading to an immediate safety stop. Nonetheless, this was not the case for our novel AIC: its strong capabilities to cope with unmodeled dynamics allowed to transfer the controller from the simulation to the real setup without re-tuning. For clarity, we only report the response of the AIC during the initial part of the pick and place cycle ($\bm q_B \to \bm q_A \to \bm q_B$) in Fig.~\ref{fig:experiments}. Joint 7 is not reported to limit redundant information, since no motion was required. As can be seen the AIC can successfully control the manipulator, however, the effect of the large uncertainties introduced for the tuning, resulted in some initial jittering, especially in joint~6\footnote{\texttt{https://youtu.be/Vsb0MzOp{\_}TY} \label{footnote}}. In other words, the AIC tuned in simulation resulted too aggressive for the real robot. This is because in simulation the AIC had to compensate also for gravity, thus a faster torque update was required. The learning rate $\kappa_a$ is the same for every joint but the jittering effect is mostly visible in joint 6. This is because in the last part of the kinematic chain, the resulting inertia acting on a joint is lower, and so it is its reluctance to changes in velocities.     
To completely remove the jittering, one can simply reduce the learning rate $\kappa_a$ to lower the torque update rate. The AIC and MRAC have been tested against large external disturbances such as a human pushing the robot during motion. AIC resulted more compliant than MRAC, showing at the same time a faster and less oscillatory disturbance rejection.
\subsubsection{Pick and place with different payloads}
\label{Sec:Experiments}
In order to use the MRAC on the real robot, a severe re-tuning of 63 parameters had to be performed, to stabilise the response due to the large unmodeled dynamics. For the AIC, $\kappa_a$ has been reduced to eliminate the jittering, as well as $\sigma_q,\ \sigma_{\dot{q}}$ to give more importance to the measurements and further reduce oscillations. The two controllers are used to perform a pick and place of an almost empty water bottle ($\approx 0.1\ [kg]$) and a full water bottle ($\approx 0.7\ [kg]$), as in Fig.~\ref{fig:pandaGazebo}. In Fig. \ref{fig:pickplacesimreal}A we show the responses of AIC and MRAC in case of the almost empty bottle, during \textit{lifting} and \textit{placing}. As can be seen, the AIC presents a faster convergence to the set-point, as well as smoother trajectories with less oscillations. To achieve a satisfactory response, we had to increase the stiffness of the MRAC, while the AIC could be kept compliant. Furthermore, in Fig. \ref{fig:pickplacesimreal}B we show the difference of the trajectories in joint space between the case with empty and full bottle, considering \textit{lifting, placing} and \textit{releasing}. Both controllers adapt to the heavier payload, making the trajectory converge to the one with lightweight bottle. AIC behaves similarly to the MRAC, yet it presents considerably less oscillations which reflected in smoother placing of the heavy object. The bigger error appearing at around $16\ [s]$ is due to the releasing of the heavy object. The effect is more visible in the AIC since the robot is more compliant. In a sense, the AIC behaves similarly to a human arm, when an unexpected weight is dropped. This is an additional evidence of the bio-inspired character of the controller. The AIC can also be tuned to be stiffer if this effect is not desired.
\subsection{Discussion and implementation notes}
Our novel AIC showed high adaptability, allowing to transfer from simulation to real robot without re-tuning. Furthermore, the AIC showed superior performance with respect to the MRAC in pick and place scenarios. The AIC is compliant while allowing to compensate for large perturbations. However, even though there is a strong evidence of stability and robustness of the AIC for a complex non-linear system, finding a formal stability proof is still an open question. Similarly to a linear case, one should determine a set of learning rates which guarantees convergence. Intuitively, active inference is a gradient descent on a quadratic and convex function thus, for some set of learning rates, the algorithm should converge to the global minima. A possible approach to a formal proof is to use Lyapunov theory as for the back-propagation algorithm in neural networks. Active inference is, in a sense, back-propagating the sensitivity of the control input with respect to the free-energy, to minimise $\mathcal{F}$. Properly addressing this proof mathematically would require a deep analysis which is out of the scope of the current paper. Another remark relates to the computational load of AIC. According to Algorithm~\ref{alg:AIC}, our novel AIC has a computational complexity of $\mathcal{O}(n)$ where $n$ is the number of DOFs. Given the structure of the generative models and covariance matrices chosen, the AIC reduces to 16 sums of vectors and 15 scalar-vector multiplications with $n$-dimensional vectors. On the other hand, the complexity of the MRAC is $\mathcal{O}(n^3)$. Another optimised computed torque algorithm such as LGP \cite{comp1}, which relies on learning dynamical models, has a cost of $\mathcal{O}(N^2)$ for online learning, where $N$ is the number of data points (i.e. $N\approx 300$). Finally, the Franka Emika Panda requires the control signals to be ready within $300\ [\mu s]$ to guarantee a functioning frequency of $1\ [kHz]$: Our AIC can perform at such a high loop rate without any package loss; is straightforward to implement; and extremely simple to tune. The source code for simulations\footnote{\texttt{https://github.com/cpezzato/panda{\_}simulation}} and experiments\footnote{\texttt{https://github.com/cpezzato/active{\_}inference}} is freely available on GitHub.

\section{Conclusion}
\label{Sec:Conclusions}
In  this paper  we  derived  the  first active inference torque controller for online joint space control of robot manipulators. Our approach makes use of the alleged adaptability of active inference, to introduce simplifications for the generative models, obtaining a model-free scheme which is less sensitive to unmodeled dynamics, is easily scalable to high DOF and is computationally inexpensive. With the proposed controller structure we overcame the complexity barrier of previous approaches, making possible control loops at high frequency with active inference. Simulations and experiments in a real setup with a 7-DOF robot arm showed that our AIC is suitable for tasks in which the dynamic model of the plant is unknown or subject to large changes. The performance of our novel AIC has been compared with that of a state-of-the-art MRAC, in different pick and place scenarios. The AIC shows better adaptability properties, allowing to transfer from simulation to real setup without re-tuning. In addition, the AIC resulted easier to tune and implement. With this work we confirmed the value of active inference to develop more adaptive control of robot manipulators. This is only the first step in this direction, future work should proof the closed-loop stability of active inference, define generative models to account for dynamic requirements and motion constraints, and be extended to other control modalities, such as control in Cartesian space or impedance control.

\section*{Acknowledgment}
The authors would like to thank Prof. Dr. Martijn Wisse for the helpful discussions, together with the whole group working on active inference at the Cognitive Robotics department.

\bibliographystyle{IEEEtran}
\bibliography{IEEEabrv,mybibfile}

\begin{thebibliography}{10}
\providecommand{\url}[1]{#1}
\csname url@samestyle\endcsname
\providecommand{\newblock}{\relax}
\providecommand{\bibinfo}[2]{#2}
\providecommand{\BIBentrySTDinterwordspacing}{\spaceskip=0pt\relax}
\providecommand{\BIBentryALTinterwordstretchfactor}{4}
\providecommand{\BIBentryALTinterwordspacing}{\spaceskip=\fontdimen2\font plus
\BIBentryALTinterwordstretchfactor\fontdimen3\font minus
  \fontdimen4\font\relax}
\providecommand{\BIBforeignlanguage}[2]{{%
\expandafter\ifx\csname l@#1\endcsname\relax
\typeout{** WARNING: IEEEtran.bst: No hyphenation pattern has been}%
\typeout{** loaded for the language `#1'. Using the pattern for}%
\typeout{** the default language instead.}%
\else
\language=\csname l@#1\endcsname
\fi
#2}}
\providecommand{\BIBdecl}{\relax}
\BIBdecl

\bibitem{NNgeneral1}
S.~Vijayakumar and S.~Shaal, ``Locally weighted projection regression:
  Incremental real time learning in high dimensional space,'' in \emph{Proc. of
  Int. Conf. on Machine Learning ({ICML})}, 2000, pp. 1079--1086.

\bibitem{NNgeneral2}
D.~Nguyen-Tuong, J.~Peters, and M.~Seeger, ``Local gaussian process regression
  for real time online model learning,'' in \emph{Proc of Neural Information
  Processing Systems ({NIPS}2008)}, 2008, pp. 1193--1200.

\bibitem{NNMatteucci}
M.~Matteucci, ``Elearnt: Evolutionary learning of rich neural network
  topologies,'' Carnegie Mellon University, 2006, technical repository No.
  CMU-CALD-02.

\bibitem{NNFOP}
F.~Ledezma and S.~Haddadin, ``First-order-principles-based constructive network
  topologies: An application to robot inverse dynamics,'' in \emph{IEEE-RAS
  17th Int. Conf. on Humanoid Robotics (Humanoids)}, 2017.

\bibitem{NNtask1}
D.~Keppler, F.~Peters, N.~Ratliff, and S.~Shaal, ``A new data source for
  inverse dynamics learning,'' in \emph{Proc of {IEEE/RJS} Conference on
  Intelligent Robots and Systems}, 2017.

\bibitem{NNtask2}
L.~Jamone, B.~Damas, and J.~Santos-Victor, ``Incremental learning of
  context-dependent dynamic internal models for robot control,'' in \emph{Proc.
  of the {IEEE} Int. Symposium on Intelligent Control ({ISIC})}, 2014.

\bibitem{friston1}
K.~J. Friston, J.~Mattout, and J.~Kilner, ``Action understanding and active
  inference,'' \emph{Biological cybernetics}, vol. 104(1-2), 2011.

\bibitem{buckley}
C.~Buckley, C.~Kim, S.~McGregor, and A.~Seth, ``The free energy principle for
  action and perception: A mathematical review,'' \emph{Journal of Mathematical
  Psychology}, vol.~81, pp. 55--79, 2017.

\bibitem{tutorial}
R.~Bogacz, ``A tutorial on the free-energy framework for modelling perception
  and learning,'' \emph{Journal of mathematical psychology}, 2015.

\bibitem{friston2}
K.~J. Friston, ``The free-energy principle: a unified brain theory?''
  \emph{Nature Reviews Neuroscience}, vol. 11(2), pp. 27--138, 2010.

\bibitem{friston3}
K.~J. Friston, J.~Daunizeau, and S.~Kiebel, ``Action and behavior: a
  free-energy formulation,'' \emph{Biological cybernetics}, vol. 102(3), 2010.

\bibitem{lopez}
L.~Pio-Lopez, A.~Nizard, K.~Friston, and G.~Pezzulo, ``Active inference and
  robot control: a case study,'' \emph{Journal of The Royal Society Interface},
  vol. 13(122), 2016.

\bibitem{arnau}
A.~C. Mercad\'e, ``Robot manipulator control under the active inference
  framework,'' \emph{(Unpublished MSc thesis), TU Delft}, 2018.

\bibitem{lanillos1}
P.~Lanillos and G.~Cheng, ``Adaptive robot body learning and estimation through
  predictive coding,'' in \emph{({IROS})}, 2018.

\bibitem{lanillos2}
P.~Lanillos and G.Cheng, ``Active inference with function learning for robot
  body perception,'' in \emph{International Workshop on Continual Unsupervised
  Sensorimotor Learning ({ICDL-Epirob})}, 2018.

\bibitem{oliver}
G.~Oliver, P.~Lanillos, and G.~Cheng, ``Active inference body perception and
  action for humanoid robots,'' arXiv:1906.03022v2, 2019.

\bibitem{astrom}
K.~Astrom, ``Theory and applications of adaptive control - a survey,''
  \emph{Automatica}, vol. Vol. 19, No. 5, pp. 471--486, 1983.

\bibitem{AdaptiveReview}
T.~Hsia, ``Adaptive control of robot manipulators - a review,'' in \emph{Proc
  of {IEEE} Int. conf. on robotics and automation ({ICRA})}, 1986.

\bibitem{MRACreview}
D.~D. Zhang and B.~Wei, ``A review on model reference adaptive control of
  robotic manipulators,'' \emph{Annual Reviews in Control}, vol.~43, 2017.

\bibitem{hyper}
M.~Tarokh, ``Hyperstability approach to the synthesis of adaptive controllers
  for robot manipulators,'' in \emph{Proc of {IEEE} international conference on
  robotics and automation ({ICRA})}, 1991.

\bibitem{selftuning1}
R.~Walters and M.~Bayoumi, ``Application of a self-tuning pole-placement
  regulator to an industrial manipulator,'' in \emph{Proc of 21st {IEEE}
  Conference on Decision and Control}, 1991, pp. 323--329.

\bibitem{selftuning2}
A.~Koivo and T.~Guo, ``Adaptive linear controller for robotic manipulators,''
  \emph{IEEE Transactions and Automatic Control}, vol. AC-28, pp. 162--171,
  1983.

\bibitem{Bayes}
D.~Lindley, ``Bayesian statistics, a review,'' \emph{SIAM}, vol.~2, 1972.

\bibitem{variational}
K.~Friston, J.~Mattout, N.~Trujillo-Barreto, J.~Ashburner, and W.~Penny,
  ``Variational free energy and the laplace approximation,'' \emph{Neuroimage},
  vol. 34(1), pp. 220--234, 2007.

\bibitem{friston5}
K.~Friston, K.~Stephan, B.~Li, and J.~Daunizeau, ``Generalised filtering,''
  \emph{Mathematical Problems in Engineering}, 2010.

\bibitem{friston4}
K.~Friston, ``Hierarchical models in the brain,'' \emph{PLoS computational
  biology}, vol. 4(11), e1000211, 2008.

\bibitem{friston6}
K.~Friston, J.~Daunizeau, and S.~Kiebel, ``Reinforcement learning or active
  inference?'' \emph{PloS one}, vol. 4(7), e6421, 2009.

\bibitem{comp1}
D.~Nguyen-Tuong and J.~Peters, ``Learning robot dynamics for computed torque
  control using local gaussian processes regression,'' in \emph{Symp. on
  Learning and Adaptive Behaviors for Robotic Systems}, 2008.

\end{thebibliography}

\end{document}